# Physics-Informed Neural Network Frameworks for the Analysis of Engineering and Biological Dynamical Systems Governed by Ordinary Differential Equations

*Andrew Particka*[*a], *Tyrus Whitman*[a], *Christopher Diers*[b], *Ian Griffin*[a], *Charuka Wickramasinghe*[c], *and Pradeep Ranaweera*[a]

[a] *Department of Engineering, Siena Heights University, Adrian, MI*
[b] *Department of Computer Information Systems, Siena Heights University, Adrian, MI*
[c] *Department of Oncology, Wayne State University, Detroit, MI*

Students: apartick@sienaheights.edu[*], twhitman@sienaheights.edu, cdiers@sienaheights.edu, igriffi1@sienaheights.edu
Mentors: gi6036@wayne.edu[c], pranawee@sienaheights.edu[a]

**ABSTRACT**
This study, presents and validate the predictive capability of the Physics-Informed Neural Networks (PINNs) methodology for solving a variety of engineering and biological dynamical systems governed by ordinary differential equations (ODEs). While traditional numerical methods are effective for many ODEs, they often struggle to achieve convergence in problems involving high stiffness, shocks, irregular domains, singular perturbations, high dimensions, or boundary discontinuities. Alternatively, PINNs offer a powerful approach for handling challenging numerical scenarios. In this study, classical ODE problems are employed as controlled testbeds to systematically evaluate the accuracy, training efficiency, and generalization capability of the PINNs framework under controlled conditions. Although not a universal solution, PINNs can achieve superior results by embedding physical laws directly into the learning process. The existence and uniqueness properties of several benchmark problems are first analyzed, and the PINNs methodology is subsequently validated on each model system. The results demonstrate that, for complex problems to converge to correct solutions, the loss function components data loss, initial condition loss, and residual loss must be appropriately balanced through careful weighting. It was further established that systematic tuning of hyperparameters including network depth, layer width, activation functions, learning rate, optimization algorithms, weight initialization schemes, and collocation point sampling plays a crucial role in achieving accurate solutions. Additionally, embedding prior knowledge and imposing hard constraints on the network architecture, without losing the generality of the ODE system, significantly enhances the predictive capability of PINNs.

**1. INTRODUCTION**
Analytical solutions to ordinary and partial differential equations (ODEs/PDEs) are often unattainable, making numerical, discretization based methods the standard alternative.[1-3] While established techniques like Euler's method and the Runge-Kutta method, for ODEs, or the finite element method and finite difference method for PDEs, have enabled the simulation of complex systems, yet they are hampered by significant limitations. Those include: the challenges of mesh generation, inefficiency with high-dimensional problems, and difficulties in resolving stiffness, complex boundary conditions, and singular perturbations.[4-7] This study consider several benchmark dynamical systems (the Lorenz system, Lotka-Volterra equation, coupled mass–spring system, and RLC circuit equation) that are usually modeled using ODEs that describe the time evolution of a quantity which needs to be solved effectively from noisy measurements or sparse data. To explore more about other interesting dynamical systems, the following references are recommended.[8-12]

Training deep neural networks for scientific applications is often hindered by a lack of large datasets. While purely data-driven models can fit existing data, they frequently produce physically implausible results and generalize poorly due to extrapolation and biases. A more robust solution is physics-informed learning, which integrates prior knowledge from physical laws to enhance model performance. PINNs have emerged as a leading method in this area, embedding differential equations directly into the learning process. By leveraging full physical knowledge and effectively adding constraints to the optimization algorithm PINNs can be trained with minimal to no labeled data to create accurate surrogate models, where the loss function quantifies the discrepancy between the model's predictions, and the physical constraints. This study presents and validates PINNs approaches with problem specific strategies, optimizing the solution to dynamical systems governed by ODEs.

In the last eight years, PINNs have been used effectively in diverse applications,[13-16] such as compartmental brain modeling, engineering, and computer vision. To obtain the approximate solution to a differential equation via deep learning, a key step is to constrain the neural network to minimize the residual (of the differential equation). Compared to the traditional mesh-based methods, deep learning could offer a mesh free approach by taking advantage of the automatic differentiation,[17] and could break the curse of dimensionality.[18] Another attractive feature of PINNs is that it can be used to solve inverse problems with minimum change of the code for forward problems.[19] In addition, PINNs have been further extended to solve integro-differential equations, fractional differential equations,[20] and stochastic differential equations.[21]

This study, demonstrate several benchmark examples using the PINNs framework. The implementation of the PINNs framework is carried out with the DeepXDE library in Python: which serves as both an educational and research platform for computational science and engineering applications. DeepXDE[22] can handle multi-physics problems and complex geometries through the constructive solid geometry approach, eliminating the need for intricate and time-consuming geometric preprocessing. Moreover, it allows users to solve time-dependent and steady-state differential equations with ease by specifying appropriate initial conditions. The framework also provides flexible callback functionalities for monitoring and customizing the training process.

The remainder of this paper is structured as follows. *Section 2* presents the general existence and uniqueness theorem for nonlinear system of ODEs and validates the theorem for the benchmark problems discussed in the study. *Section 3* outlines the PINNs methodology employed to solve these ODEs. *Section 4* provides numerical examples that demonstrate the predictive capability of the PINNs approach. Finally, *Section 5* concludes the results and highlights potential directions for future research.

## 2. METHODS AND PROCEDURES

Before addressing methods for solving differential equations: whether analytically, qualitatively, or numerically, it is essential to first examine the fundamental mathematical questions of existence and uniqueness. The first concern is whether a solution exists at all; without existence, the search for a solution is meaningless. The second concern is uniqueness: if multiple solutions arise from the same problem formulation, the equation loses its predictive power and may have limited applicability to physical systems. Since differential equations often admit infinitely many solutions, uniqueness can only be guaranteed by imposing appropriate initial or boundary conditions. Thus, as a preliminary step, a set of theorems and conditions is stated to provide the theoretical foundation for verifying the existence and uniqueness of solutions to the ODE systems presented in this study to explore the predictive power of the PINNs framework. The proof of the theorems can be found from.[23]

### 2.1. Existence and Uniqueness

**Theorem 1** (Existence and Uniqueness Theorem for Linear Equations) Let $p_1(t), \ldots, p_{n-1}(t)$ and $g(t)$ are continuous on an interval $(a, b)$ containing the point $t_0$. Then for every choice of initial values $y_0, y_1, \ldots, y_{n-1}$ there exists a unique solution on the whole interval $(a, b)$ to the initial value problem.

$$y^{(n)}(t) + p_1(t) y^{(n-1)}(t) + \cdots + +p_n(t) y(t) = g(t);$$
$$y(t_0) = y_0, y'(t_0) = y_1, \ldots, \quad y^{(n-1)}(t_0) = y_{n-1}.$$

**Theorem 2** (Picard–Lindelöf Existence and Uniqueness Theorem for Systems) If $\boldsymbol{f}$ and $\frac{\partial \boldsymbol{f}}{\partial x_i}, i = 1,2,\ldots,n$, are continuous function in a 'rectangle $R = \{(t, x_1, \ldots, x_n): \ a < t < b, \ c_i < x_i < d_i, \ i = 1, \ldots, n\}$ that contains the point $(t_0, \boldsymbol{x_0})$, then the initial values problem $\boldsymbol{x'}(t) = \boldsymbol{f}(t, \boldsymbol{x})$, $\boldsymbol{x}(t_0) = \boldsymbol{x_0}$, where $\boldsymbol{x_0} = col(x_{1,0}, \ldots, x_{n,0})$, has a unique solution in some interval $[t_0 - h, t_0 + h]$, $h$ is a positive constant.

**Theorem 3** (Continuation of Solution) Let $\boldsymbol{x} = (x_1, \ldots, x_n)$ and $\boldsymbol{f}(t, \boldsymbol{x})$ denote the vector function $\boldsymbol{f}(t, \boldsymbol{x}) = (f_1(t, x_1, \ldots, x_n), \ldots, f_n(t, x_1, \ldots, x_n))$. Suppose $\boldsymbol{f}$ and $\frac{\partial \boldsymbol{f}}{\partial x_i}, i = 1,2,\ldots,n$, are continuous on the strip $R = \{(t, \boldsymbol{x}): \ a \leq t \leq b, \ \boldsymbol{x} \ arbitrary\}$ containing the point $(t_0, \boldsymbol{x_0})$. Assume further that there exist a positive constant $L$ such that for $i = 1,2, \ldots, n$, $\left|\frac{\partial f}{\partial x_i}\right| \leq L$ for all $t, \boldsymbol{x})$ in $\boldsymbol{R}$. Then the initial value problem $\boldsymbol{x'}(t) = \boldsymbol{f}(t, \boldsymbol{x})$, $\boldsymbol{x}(t_0) = \boldsymbol{x_0}$, has a unique solution on the entire interval $a \leq t \leq b$.

**Lemma 1** (Grönwall's inequality) Let *I* denote an interval of the real line of the form $[a, b]$ with $a < b$. Let $\beta$ and *u* be real valued continuous functions defined on I. If u is differentiable in the interior of I and satisfies the differential inequality $u'(t) \leq$ $\beta(t)u(t)$, $t \in (a,b)$ then $u$ is bounded by the solution of the corresponding differential equation $v'(t) \leq \beta(t)v(t)$: $u(t) \leq u(a)\exp\left(\int_a^t \beta(s)ds\right)$ for all, $t \in I$.

### 2.1.1. The Lorenz System

The Lorenz system is a set of three ordinary differential equations, first developed by the meteorologist Edward Lorenz while studying atmospheric convection. It is a classic example of a system that can exhibit chaotic behavior, meaning its output can be highly sensitive to small changes in its starting conditions. For certain values of its parameters, the system's solutions form a complex, looping pattern known as the Lorenz attractor as shown in **Figure 1** The model describes how three key properties of this system change over time as follows:

$$\frac{dx}{dt} = \delta(y - x)$$
$$\frac{dy}{dt} = x(\rho - z) - y$$
$$\frac{dz}{dt} = xy - \beta z \qquad t \in [a,b]$$

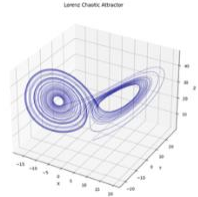

**Figure 1.** The Lorenz attractor.

with initial conditions $x(0) = x_0$, $y(0) = y_0$, $z(0) = z_0$, where, $x$ is proportional to the intensity of the convection (the rate of fluid flow), $y$ is proportional to the temperature difference between the rising and falling air currents and $z$ is proportional to the distortion of the vertical temperature profile from a linear one. To verify the existence and the uniqueness of the solution, let $f(t,x) = \begin{bmatrix} \delta(y-x) \\ x(\rho-z)-y \\ xy-\beta z \end{bmatrix}$ and thus $\frac{\partial f}{\partial x} = \begin{bmatrix} -\delta & \delta & 0 \\ \delta-z & -1 & -x \\ y & x & -\beta \end{bmatrix}$. Each component of $f(t,x)$ is a polynomial in $(x,y,z)$ and thus continuous as well as all the entries of $\frac{\partial f}{\partial x}$ are continuous. Then by the Picard–Lindelöf theorem: $f$ is locally Lipschitz, hence for any initial condition $\mathbf{x}(t_0) = \mathbf{x}_0$ there exists a unique local solution on some interval $t \in [t_0 - \epsilon, t_0 + \epsilon]$. To extend the solution globally, define the energy-like function $V(x,y,z) = x^2 + y^2 + (z-\rho)^2 \geq 0$. Differentiating along solutions: $\dot{V} = 2x\dot{x} + 2y\dot{y} + 2(z-\rho)\dot{z}$. The we substitute the Lorenz equations to $\dot{V}$ and obtain:

$$\dot{V} = 2x\sigma(y-x) + 2y(x(\rho-z)-y) + 2(z-\rho)(xy-\beta z)$$

After expansion and applying standard inequalities ($2ab \leq a^2 + b^2$) and completing the square for the $z$-term:

$$\dot{V} \leq KV + C_0$$

where $K > 0$ and $C_0 > 0$ are constants depending on $\sigma, \rho, \beta$. Then by Grönwall's inequality, it follows that, $V(x,y,z)$ and therefore $(x(t), y(t), z(t))$, remains bounded for all $t \geq t_0$.

### 2.1.2. Lotka -Volterra Equation

The Lotka–Volterra equations, also known as the Lotka–Volterra predator–prey model, are a pair of first-order nonlinear differential equations, frequently used to describe the dynamics of biological systems in which two species interact, one as a predator and the other as prey. The populations change through time according to the pair of equations:

$$\frac{dx}{dt} = \alpha x - \beta xy, \quad \frac{dy}{dt} = -\gamma y + \delta xy;$$ with initial conditions $x(0) = x_0$, $y(0) = y_0$.

To verify the existence and the uniqueness of the solution, let $f(t,x) = \begin{bmatrix} \alpha x - \beta xy \\ -\gamma y + \delta xy \end{bmatrix}$ and thus $\frac{\partial f}{\partial x} = \begin{bmatrix} \alpha - \beta y & -\beta x \\ -\delta y & -\gamma + \delta x \end{bmatrix}$. Each component of $f(t,x)$ is a polynomial in $(x,y,z)$ and thus continuous as well as all the entries of $\frac{\partial f}{\partial x}$ are continuous. Then by the Picard–Lindelöf theorem: $f$ is locally Lipschitz, hence for any initial condition $\mathbf{x}(t_0) = \mathbf{x}_0$ there exists a unique local solution on some interval $t \in [t_0 - \epsilon, t_0 + \epsilon]$.

### 2.1.3. Coupled Mass - Spring System

A coupled mass–spring system is considered, consisting of two masses connected by springs, where the motion of each mass affects the other. The governing equation for the mass–spring system can be written as:

$$[inretia]x'' + [damping]x' + [stiffness]x = F_{external}$$

The variables $x$ and $y$ typically represent the displacements of two masses from their equilibrium positions. The primes $(x'', y'')$ represent acceleration of mass 1, and mass 2 respectively. **Figure 2** shows on a smooth horizontal surface, a mass is attached to a fixed wall by a spring, and another mass is attached to the first object by a spring. The objects are aligned horizontally so that the springs are their natural lengths. In this study, $F_{external} = 0$ is set, and the coupled mass–spring system is defined by the following system of second-order linear differential equations:

$$m_1 x'' + (k_1 + k_2)x - k_2 y = 0 \qquad\qquad m_2 y'' + (k_2 y - k_2)x = 0$$

For simplicity set, $m_1 = 2$, $m_2 = 1$, $k_1 = 1$, and $k_2 = 1$ with $x(0) = 3$, $x'(0) = 0$ $y(0) = 3$, and $y'(0) = 0$ to discuss the existence uniqueness. To this end, first convert Mass–Spring System into a system of first order differential equation using the transformations $u_1 = x$, $u_2 = x'$, $u_3 = y$ and $u_4 = y'$. Then obtain the order reduction system as follows:

$$u' = f(t, u) = \begin{bmatrix} u_2 \\ u_3 - 3u_1 \\ u_4 \\ 2u_1 - 2u_3 \end{bmatrix}, \qquad u(0) = u_0 = \begin{bmatrix} 3 \\ 0 \\ 3 \\ 0 \end{bmatrix}.$$

Since $f(t, u)$ is linear in $u$ and independent of $t$, $f(t, u)$ is globally Lipschitz. Therefore, by Picard–Lindelöf theorem there exists a unique global solution for all $t$. Additionally, consider the linear system in matrix form:

$$u' = Au, \quad u = \begin{bmatrix} u_1 \\ u_2 \\ u_3 \\ u_4 \end{bmatrix}, \quad A = \begin{bmatrix} 0 & 1 & 0 & 0 \\ -3 & 0 & 1 & 0 \\ 0 & 0 & 0 & 1 \\ 2 & 0 & -2 & 0 \end{bmatrix}.$$

then the eigen values of $A$ are $\lambda_1 = +2i$, $\lambda_2 = -2i$, $\lambda_2 = +i$ and $\lambda_4 = -i$. Since all eigenvalues are purely imaginary, the solution is oscillatory, which will be demonstrated in *Section 3*.

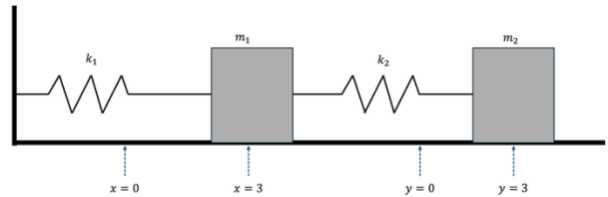

**Figure 2.** The coupled mass–spring system.

### 2.1.4. RLC Circuit Equation

An RLC circuit is an electrical circuit consisting of a resistor (R), an inductor (L), and a capacitor (C), connected in series or in parallel. The equation governing the voltage $v(t)$ across the capacitor were given as follows:

$$C \frac{d^2 v}{dt} + \frac{1}{R}\frac{dv}{dt} + \frac{1}{L}v = f(t) \quad v(0) = v_0 \;\; v'(0) = v_0'$$

For constants, C, R ($\neq 0$), and L ($\neq 0$) with a continuous function $f(t)$ RLC circuit equation has a unique solution by **Theorem 1**. If $f(t) = 0$ for $t > 0$ RLC circuit is in free oscillation meaning how the circuit behaves after it's energized and then left to oscillate or decay with no external forcing function. The oscillation is called underdamped if $R > \sqrt{L/4C}$, overdamped if $R < \sqrt{L/4C}$, and critically damped if $R = \sqrt{L/4C}$; an example for underdamping is demonstrated in *Section 3*. **Figure 3** exhibits a parallel RLC circuit.

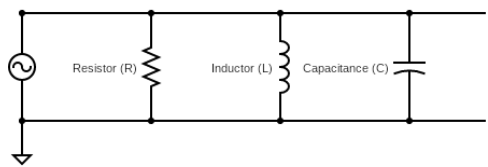

**Figure 3.** A parallel RLC circuit.

## 2.2 Physics Informed Neural Networks (PINNs)

In this section, a brief overview of deep neural networks is provided, along with the procedure of PINNs for solving differential equations. Mathematically, a deep neural network is a particular choice of compositional function. The simplest neural network is the feed forward neural network (FNN), also called multilayer perceptron (MLP), which applies linear and nonlinear transformations to the inputs recursively. Although many different types of neural networks have been developed in the past decades, this paper consider FNN, which is sufficient for most differential equation problems. Let $\mathcal{N}^L(x): \mathbb{R}^{d_{in}} \to \mathbb{R}^{d_{out}}$ be a $L$-layer neural network, with $N_l$ neurons in the $l$-th layer. The weight matrix and bias vector in the $l$-th layer is denoted by $\boldsymbol{W}^l \in \mathbb{R}^{N_l \times N_{l-1}}$ and $b^l \in \mathbb{R}^{N_l}$, respectively. Given a nonlinear activation function σ, which is applied element-wisely, the FNN is recursively defined as follows:

- Input layer: $\mathcal{N}^0(x) = x \in \mathbb{R}^{d_{in}}$;
- Hidden layers: $\mathcal{N}^l(x) = \sigma(\boldsymbol{W}^l \mathcal{N}^{l-1}(x) + \boldsymbol{b}^l) \in \mathbb{R}^{N_l}$, for $1 \leq l \leq L-1$;
- Output layer: $\mathcal{N}^L(x) = \boldsymbol{W}^L \mathcal{N}^{L-1}(x) + \boldsymbol{b}^l \in \mathbb{R}^{d_{out}}$

Commonly used activation functions including the sigmoid ($1/(1 + e^{-x})$), the hyperbolic tangent (tanh), sine ($\sin(x)$), and the rectified linear unit (ReLU, $max(x, 0)$) are shown in **Figure 4**. To measure the discrepancy between the neural network solution $\hat{u}$ and the constraints, consider the loss function defined as the weighted summation of the $L_2$ norm of residuals for the equation and boundary/initial conditions and the experimental data. The composite PINNs loss function was defined as:

$$\mathcal{L}(\boldsymbol{\theta}) = \lambda_{data}\mathcal{L}(\boldsymbol{\theta})_{data} + \lambda_{ode}\mathcal{L}(\boldsymbol{\theta})_{ode} + \lambda_{ic}\mathcal{L}(\boldsymbol{\theta})_{ic}; \text{ where,}$$

$$\mathcal{L}(\boldsymbol{\theta})_{data} = \sum_{j=1}^{m} \lambda_{data}^j \mathcal{L}_{data}^j = \sum_{j=1}^{m} \lambda_{data}^j \left[\frac{1}{n}\sum_{i=1}^{n}\left(x^j{}_{obs}(t_i) - x^j{}_{pred}(t_i; \boldsymbol{\theta})\right)^2\right],$$

$$\mathcal{L}(\boldsymbol{\theta})_{ode} = \sum_{j=1}^{s} \lambda_{ode}^j \mathcal{L}_{ode}^j = \sum_{j=1}^{s} \lambda_{data}^j \left[\frac{1}{n}\sum_{i=1}^{n}\left(\frac{d\widehat{x_j}}{dt}\big|_{t_i} - f_j\left(\hat{x}_j(t_i)\right)\right)^2\right],$$

$$\mathcal{L}(\boldsymbol{\theta})_{ic} = \sum_{j=1}^{s} \lambda_{ic}^j \mathcal{L}_{ic}^j = \sum_{j=1}^{s} \lambda_{ic}^j \left[\frac{1}{s}\left(x^j{}_{obs}(t_0) - x^j{}_{pred}(t_0; \boldsymbol{\theta})\right)^2\right].$$

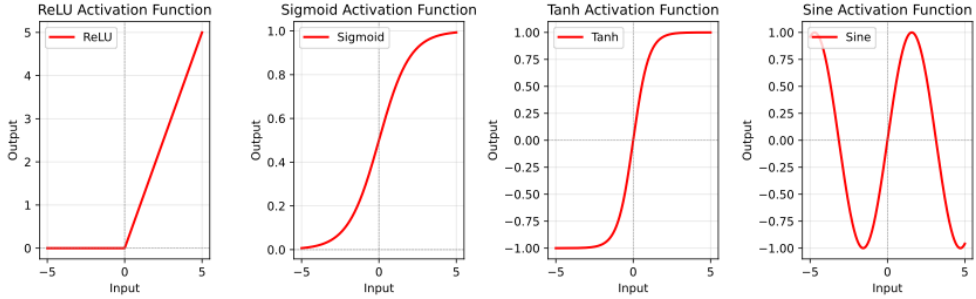

**Figure 4.** Commonly used activation functions.

The loss function in PINNs is designed to enforce both data fidelity and physical consistency by combining multiple objective terms. The data loss ($\mathcal{L}_{data}$) ensures that the neural network's predictions align with observed measurements, minimizing discrepancies at known data points. Data loss is calculated by the sum of the square differences of the predicted concentrations and the observed data. The ODE loss ($\mathcal{L}_{ODE}$), embedding the underlying dynamics directly into the learning process makes sure to fulfill the law described by ODEs. The initial condition loss ($\mathcal{L}_{IC}$), guarantees that the solution adheres to prescribed initial constraints. weights $\lambda_{data}$, $\lambda_{ODE}$, $\lambda_{IC}$ balance contributions from data fidelity, ODE residuals, and initial conditions. The derivatives $\frac{d\widehat{x_j}}{dt}\big|_{t_i}$ are analytically calculated with automatic differentiation. By optimizing the composite loss function ($\mathcal{L}(t; \boldsymbol{\theta})$),

$$\boldsymbol{\theta}^* = argmin[\mathcal{L}(t; \boldsymbol{\theta})],$$

the network not only interpolates sparse data but also generalizes as a physics-compliant surrogate model that is robust in regions where measurements are unavailable. **Figure 5** illustrates the PINN framework, where time (t) serves as input and outputs are optimized to minimize total loss. During the minimization we infer the neural network parameters *θ* (weights and bias) via gradient-based optimizers, such as Adam and L-BFGS.[24,25] The convergence and accuracy of the PINNs are critically dependent on the selection of its architecture (*i.e.* number of layers and the neurons), optimizer (Adam or L-BFGS), learning rate ($\eta$), weights (*w*) and bias (*b*) initializers (Glorot normal, Glorot uniform), number of optimization iterations, type of activation function (such as Tanh, Sigmoid, ReLU, Swish, Sine, *etc.*), and loss weights ($\lambda_{data}$, $\lambda_{ode}$, $\lambda_{ic}$). In this study, the loss weight coefficients are manually selected according to the problem of interest, ensuring that the weighted losses remain of the same order of magnitude during network training. An open issue is that PINNs can produce negative values, even for strictly non-negative quantities. To handle this issue, a transformation is defined to ensure that the output is positive during training. If the nature of the solution curves is known, this information is incorporated into the network by adding a feature layer, enhancing accuracy and producing biologically meaningful solutions. The physics-informed neural network procedure for solving ODEs through the DeepXDE library is presented in the following.

*PINNs procedure for solving differential equations through DeepXDE*

| | |
|---|---|
| Step 1 | Define the computational domain using the **geometry** module. |
| Step 2 | Define the differential equation. |
| Step 3 | Define the boundary and initial conditions. |
| Step 4 | Combine geometry, differential equations, and boundary/initial conditions together into **data.PDE.** |
| Step 5 | Specify the network architecture through layers, neurons, activation functions, inputs, outputs and wight initializers into **nn.FNN.** |
| Step 6 | Combine **data.PDE** and the **nn.FNN** into dde.Model. |
| Step 7 | Call **Model.compile** to set the optimization hyperparameters, such as optimizer and learning rate. The weights can be set here by loss weights. |
| Step 8 | Call **Model.train** to train the network where the number of iterations can be specified. |
| Step 9 | Call **Model.predict** to predict the solution at user defined, time points. |

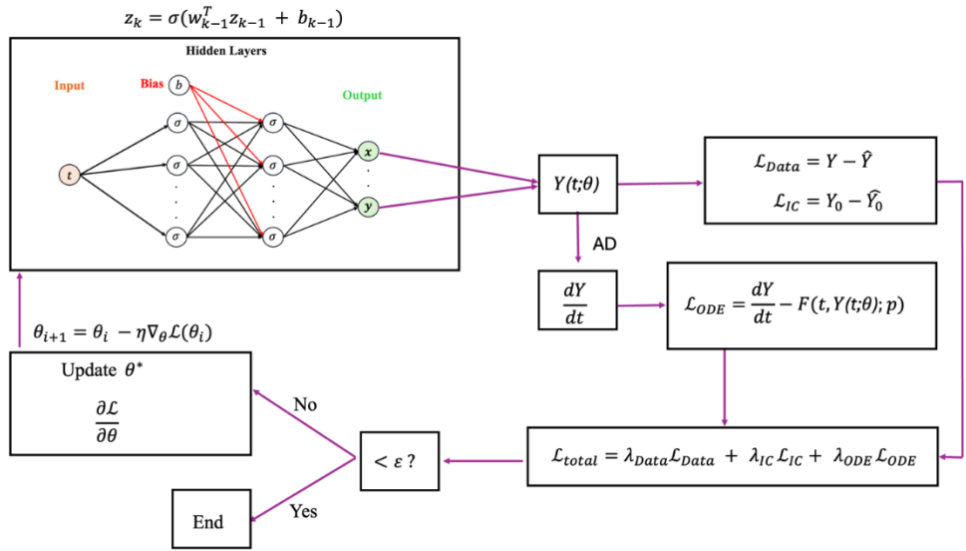

**Figure 5.** The physics-informed neural network model starts with time as inputs and then the outputs go to an optimization block where it minimizes the total loss $L_{Total}$ by optimizing the parameters θ via gradient-based optimizers.

**RESULTS AND DISCUSSION**

In this section, several numerical examples are presented to evaluate the predictive capabilities of PINNs. In Example 1, the influence of network architecture, activation functions, and loss weights on the solution is investigated. *Example 2* illustrates the ability of PINNs to approximate solutions when additional features are incorporated. *Example 3* is designed to assess prediction accuracy in the presence of noisy data. Finally, *Example 4* demonstrates the performance of PINNs in scenarios without experimental data. The total loss is computed as described in *Theorem 2*. For a function $u(x)$ (the true solution) and its PINN approximation $\hat{u}(x)$, the relative $L_2$ error is calculated at $N$ discrete points $x_i$ as follows:

$$L_2 \; error = \frac{\sqrt{\sum_{i=1}^{N}(\hat{u}(x_i)-u(x_i))^2}}{\sqrt{\sum_{i=1}^{N}(u(x_i))^2}}.$$

**Example 1.** $L_2$ *Error and Computational Efficiency*

In this example, the performance of PINNs is evaluated under varying network configurations, including different numbers of layers, neurons, and activation functions. The significance of appropriate weighting through components of the loss function is further emphasized. For each simulation, the $L_2$ error and computational time are recorded. To this end, the RLC circuit is considered and is given by:

$$c\frac{d^2v}{dt} + \frac{1}{R}\frac{dv}{dt} + \frac{1}{L}v = f(t) \quad v(0) = v_0 \quad v'(0) = v'_0$$

where, $R = 20000$, $L = 8$, $C = 0.125\ e^{-6}$, and $f(t) = 0$. **Table 1** provides a detailed summary of the simulation results. All simulations presented in **Table 1** were conducted using the Adam optimizer with a learning rate of 0.0001. The network weights were initialized using the Glorot normal initializer, and each model was trained for 100,000 epochs. Due to the singularly perturbed nature of the *RLC Circuit Equation*, the PINNs failed to accurately capture the true solution when trained with arbitrary hyperparameters. Therefore, a systematic analysis was performed, as summarized in **Table 1**. The results indicate that the PINNs successfully captured the true solution when configured with 3 layers, 25 neurons per layer, sine activation function, and loss weights of $[10^{-7},\ 10^{+3},\ 10^0,\ 10^0]$. The $L_2$ error and the corresponding computational time (in seconds) are summarized in **Table 1**. The minimum $L_2$ error achieved was 0.0012, obtained after 114.7 seconds of training, which corresponds to the entry *0.0012 (114.7)* in **Table 1**. After determining the optimal configuration for obtaining an acceptable solution based on **Table 1**, we conducted simulations under identical settings while varying the number of training iterations. The corresponding results are summarized in **Table 2**. As shown, the accuracy of the predicted solution improves with an increased number of iterations; however, this improvement comes at the expense of significantly higher computational cost. **Figure 6** illustrates the network architecture, the comparison between the predicted and reference solutions, the individual components of the loss function, and the evolution of the $L_2$ error after one million iterations.

| L | AF | | Number of Neurons | | | |
|---|---|---|---|---|---|---|
| | | Loss Weights | N=5 | N=25 | N=75 | N=100 |
| 1 | Relu | $[10^0,\ 10^0,\ 10^0,\ 10^0]$ | 1.0(42.8) | 1.0(42.8) | 1.0(43.3) | 1.0(48.7) |
| | | $[10^{-7},\ 10^{+3},\ 10^0,\ 10^0]$ | 0.9999(42.8) | 0.9999(43.5) | 0.9999(47.3) | 0.9999(48.7) |
| | Tanh | $[10^0,\ 10^0,\ 10^0,\ 10^0]$ | 1.0(46.3) | 1.0(61.8) | 1.0(61.8) | 1.0(69.3) |
| | | $[10^{-7},\ 10^{+3},\ 10^0,\ 10^0]$ | 1.0295(48.3) | 1.1669(61.7) | 1.6172(69) | 1.875(72.5) |
| | Sigmoid | $[10^0,\ 10^0,\ 10^0,\ 10^0]$ | 1.0(46.5) | 1.0(48.5) | 1.0(53.9) | 1.0(56.3) |
| | | $[10^{-7},\ 10^{+3},\ 10^0,\ 10^0]$ | 1.0069(47.6) | 1.0307(48.1) | 1.1188(53.5) | 1.1671(56.3) |
| | Sine | $[10^0,\ 10^0,\ 10^0,\ 10^0]$ | 1.0(46.5) | 1.0(65.7) | 1.0(78.8) | 1.0(89.8) |
| | | $[10^{-7},\ 10^{+3},\ 10^0,\ 10^0]$ | 1.0274(48) | 1.1679(67.4) | 1.6197(82.4) | 1.8791(89.2) |
| 3 | Relu | $[10^0,\ 10^0,\ 10^0,\ 10^0]$ | 1.0(64.6) | 1.0(71.9) | 1.0(106) | 0.9999(120.8) |
| | | $[10^{-7},\ 10^{+3},\ 10^0,\ 10^0]$ | 0.9999(64.1) | 0.9999(70.8) | 0.9999(101.2) | 0.9999(117.3) |
| | Tanh | $[10^0,\ 10^0,\ 10^0,\ 10^0]$ | 1.0(75.8) | 1.0(109) | 0.9999(137.3) | 0.9999(137.3) |
| | | $[10^{-7},\ 10^{+3},\ 10^0,\ 10^0]$ | 0.9974(75.9) | 0.9964(104) | 0.9675(145) | 0.9008(153.3) |
| | Sigmoid | $[10^0,\ 10^0,\ 10^0,\ 10^0]$ | 1.0(80) | 1.0(86.9) | 1.0(150.6) | 1.0(187.2) |
| | | $[10^{-7},\ 10^{+3},\ 10^0,\ 10^0]$ | 1.0336(78.3) | 1.1947(86.4) | 1.4457(144.4) | 1.374(189.1) |
| | Sine | $[10^0,\ 10^0,\ 10^0,\ 10^0]$ | 1.0(77.5) | 1.0(122.4) | 1.0(176.7) | 1.0(288.9) |
| | | $[10^{-7},\ 10^{+3},\ 10^0,\ 10^0]$ | 0.5275(79.8) | *0.0012(114.7)* | 0.0128(172.1) | 0.0053(184.0) |
| 9 | Relu | $[10^0,\ 10^0,\ 10^0,\ 10^0]$ | 1.0(128.5) | 1.0(142.1) | 1.0(256.9) | 1.0(290.2) |
| | | $[10^{-7},\ 10^{+3},\ 10^0,\ 10^0]$ | 0.9999(129.7) | 0.9999(147.6) | 0.9999(247.5) | 0.9999(287.7) |
| | Tanh | $[10^0,\ 10^0,\ 10^0,\ 10^0]$ | 0.9442(159.8) | 0.8923(241) | 0.9992(375.8) | 1.0(817.3) |
| | | $[10^{-7},\ 10^{+3},\ 10^0,\ 10^0]$ | 0.9936(158.5) | 0.9888(233.1) | 0.9775(350.6) | 0.9841(600.1) |
| | Sigmoid | $[10^0,\ 10^0,\ 10^0,\ 10^0]$ | 1.0(165) | 1.0(196.1) | 1.0(402.7) | 1.0(534.4) |
| | | $[10^{-7},\ 10^{+3},\ 10^0,\ 10^0]$ | 0.9939(164.4) | 0.9999(194.3) | 1.0(381.9) | 1.0001(533.8) |
| | Sine | $[10^0,\ 10^0,\ 10^0,\ 10^0]$ | 0.9998(24222.2) | 0.7753(260.5) | 1.0(470.8) | 1.0(519) |
| | | $[10^{-7},\ 10^{+3},\ 10^0,\ 10^0]$ | 1.0013(8298.2) | 0.7641(4177.7) | 0.3879(481.6) | 0.6166(525.7) |

**Table 1.** A systematic model simulation approach to capture the true solution. Abbreviations: L the number of layers; N the number of neurons; AF activation function.

| Number of iterations | 5000 | 10000 | 100000 | 300000 | 1000000 |
|---|---|---|---|---|---|
| $L_2$ error | 1.12 | 1.23 | 0.0012 | 8.44e-04 | 6.19e-04 |
| Computational time (s) | 5.17 | 10.05 | 114.7 | 338.05 | 1178.00 |

**Table 2.** $L_2$ error and the computational time over iterations.

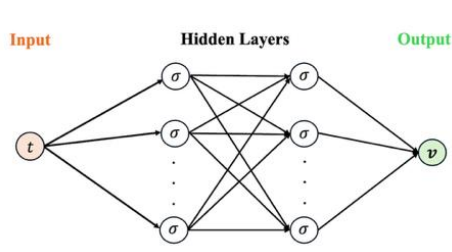
(a) The network architecture

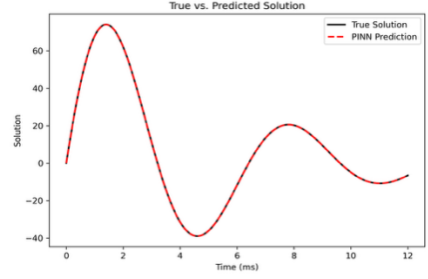
(b) The network architecture

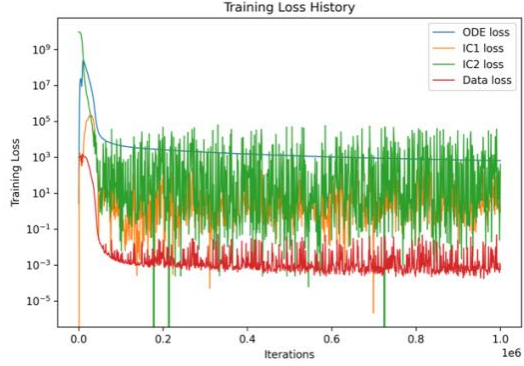
(c) Training loss history

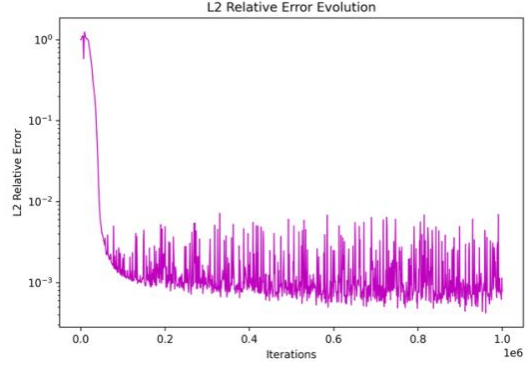
(d) L2 relative error evolution

**Figure 6.** Graphical illustration of the network, solution and the loss and error histories of *RLC Circuit Equation*.

**Example 2.** *Enhance learning by training with additional features*
The main objective of this example is to demonstrate how incorporating feature layers can enhance the performance of PINNs. Using the Lotka–Volterra Equation as an example, it is shown that constructing feature layers based on characteristic patterns observed in dynamical systems can significantly improve the network's training efficiency and accuracy. To this end, the Lotka–Volterra Equation is defined as follows:

$$\frac{dx}{dt} = 10x(1.5 - 9.5y) \qquad \frac{dy}{dt} = 10y(5.7x - 1.05)$$
$$\text{with initial conditions } x(0) = 0.5 \qquad y(0) = 0.075$$

As shown in **Figure 7(a)**, in addition to the standard layers, extra layers are added to make the network training easier. Since periodic behavior is expected in the *Lotka-Volterra Equation*, a feature layer with $f_k(t) = \sin(kt)$ where $k = 1,2,\ldots n$ is added. This approach enforces periodicity in predictions, thereby enhancing accuracy. Additionally, hard constraints are applied to the boundary conditions, ensuring that the network does not need to infer them solely through training. As shown in **Table 3**, under identical conditions, the neural network achieves improved learning performance when feature layers are incorporated. **Figure 7(b)** presents the predicted solution without feature layers alongside the exact solution, whereas **Figure 7(c)** shows the predicted solution with feature layers compared to the exact solution. Both **Table 3** and **Figure 7** demonstrate that the inclusion of feature layers can significantly enhance the performance of PINNs.

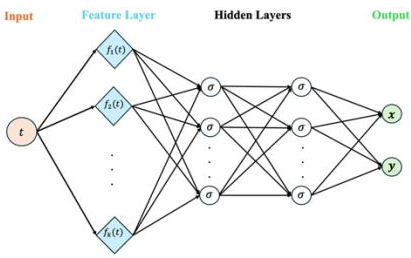 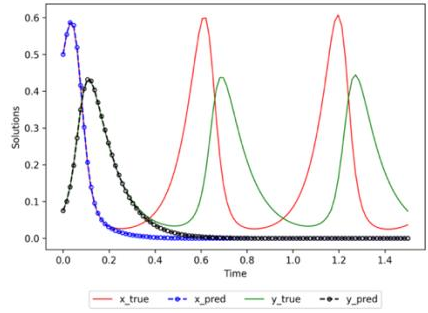 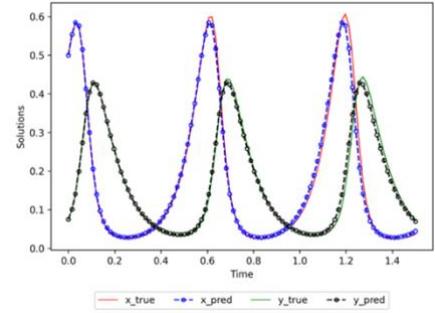

**(a)** The network architecture   **(b)** Without feature layer   **(c)** With feature layer

**Figure 7.** PINN solution to Lorenz system with noise data.

|  | L | N | AF | Alg | R | C | E | LW | $\mathcal{L}(\boldsymbol{\theta})_{ode}$ | CT |
|---|---|---|---|---|---|---|---|---|---|---|
| Without features | 6 | 64 | sine | Adams, L-BFGS | 0.001 | 400 | 20000 | [1;1;1] | 3.48e-03 | 599.81 |
| With features | 6 | 64 | sine | Adams, L-BFGS | 0.001 | 400 | 20000 | [1;1;1] | 1.10e-06 | 656.76 |

**Table 3.** The optimal network parameters NN Architecture. Abbreviations: L the number of layers; N the number of neurons; AF activation function; Alg optimization algorithm; R learning rate; C number of collocation points; E number of epochs; LW loss weights; CT computational time in seconds.

**Example 3.** *Model Performance in the Presence of Noisy Data*

This example, investigate the predictive capability of PINNs for the Lorenz system under noisy observations. The Lorenz system is defined as:

$$\frac{dx}{dt} = \delta(y - x) \quad \frac{dy}{dt} = x(\rho - z) - y \quad \frac{dz}{dt} = xy - \beta z \quad t \in [a, b]$$

with initial conditions $x(0) = x_0$, $y(0) = y_0$, $z(0) = z_0$. A reference solution was obtained using the classical numerical solver Tsit5() in Julia. To assess robustness to measurement noise, the reference solutions were contaminated with additive Gaussian noise, simulating observational errors. For each state variable $j \in \{x, y, z\}$ and each time point $t_i$, the noisy observations were generated as $\hat{x}_j(t_i) = x_j(t_i) + \sigma \cdot \varepsilon_j(t_i)$ where $\varepsilon_j(t_i) \sim \mathcal{N}(0,1)$ are independent standard normal random variables. Three noise levels ($\sigma$ = 0.2, 1.0, 3.0) were considered, corresponding to low, medium, and high noise conditions, respectively. Random seed fixation ensured reproducible noise realizations. The system parameters were set to $\delta = 10$, $\rho = 15$, and $\beta = 8/3$ with initial conditions $x_0 = -8$, $y_0 = 7$, and $z_0 = 27$. Optimal PINN hyperparameters for each noise level are summarized in **Table 4**. As noise levels increase, it may be necessary to adjust the network architecture and training parameters to preserve prediction accuracy. To this end, the number of layers, neurons, and collocation points is increased to enhance performance; however, this also results in an increase in both total loss and computational time, as shown in **Table 4**. **Figure 8** demonstrates that PINNs can accurately recover the system dynamics from noisy observations. Each simulation employed a two-stage training procedure: an initial optimization using the Adams algorithm, followed by refinement with the Limited-memory Broyden–Fletcher–Goldfarb–Shanno (L-BFGS) algorithm. **Figure 9** visualizes the resulting phase-space trajectories, illustrating the coupled evolution of $x$, $y$, and $z$ and providing a comprehensive depiction of the system's dynamics.

| σ | L | N | AF | Alg | R | C | E | LW | Loss | CT |
|---|---|---|---|---|---|---|---|---|---|---|
| Low | 3 | 40 | tanh | Adams, L-BFGS | 0.001 | 400 | 50000 | [1;1;1] | 1.24e-01 | 111.59 |
| Medium | 4 | 50 | tanh | Adams, L-BFGS | 0.001 | 500 | 75000 | [1;1;1] | 2.67e+00 | 227.64 |
| High | 5 | 60 | tanh | Adams, L-BFGS | 0.001 | 600 | 100000 | [1;1;1] | 2.40e+01 | 580.80 |

**Table 4.** The network settings for each noise level. Abbreviations: σ the noise level; L the number of layers; N the number of neurons; AF activation function; Alg optimization algorithm; R learning rate; C number of collocation points; E number of epochs; LW loss weights; CT computational time in seconds.

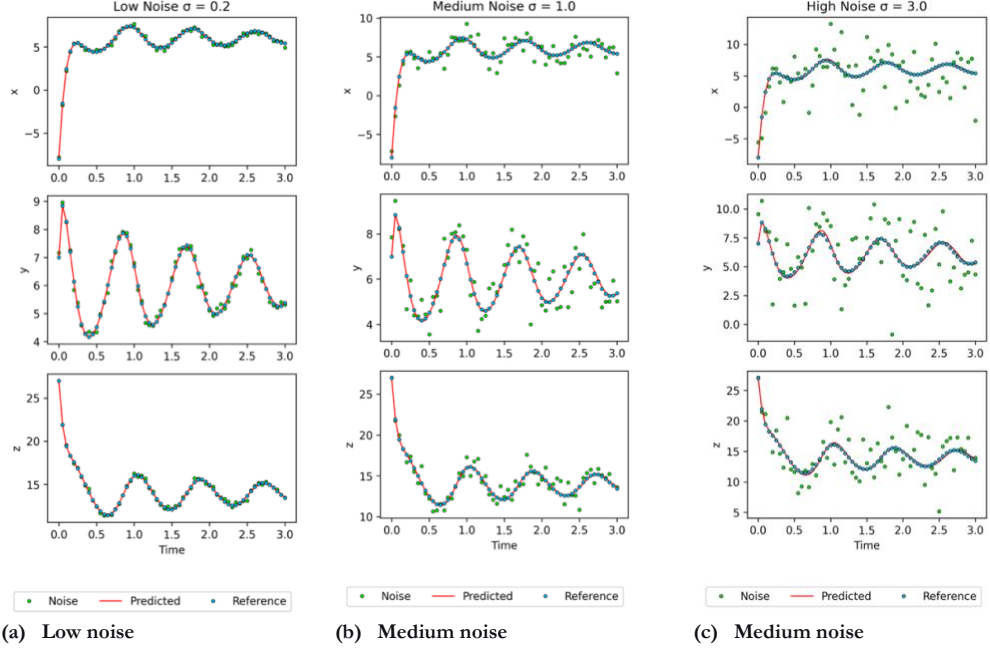

(a) Low noise    (b) Medium noise    (c) Medium noise

**Figure 8.** PINN solution to Lorenz system with noisy data.

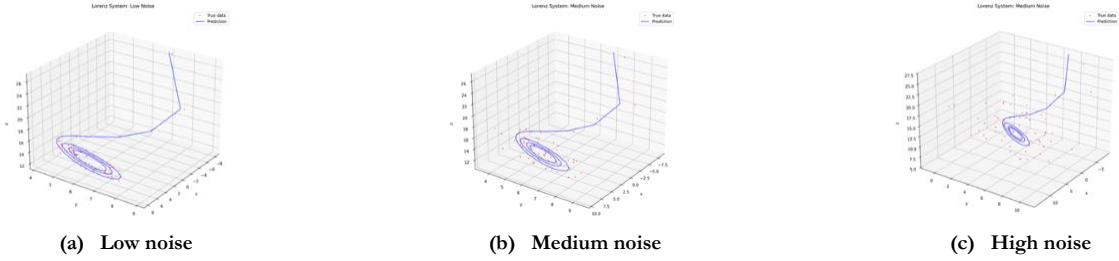

(a) Low noise    (b) Medium noise    (c) High noise

**Figure 9.** Phase plane of PINNs solutions to Lorenz system with noisy data.

***Example 4.*** *Performance in the Absence of Experimental Data*
This example, demonstrate the performance of the neural network in the absence of experimental data. Consequently, the loss function comprises only two components, as described below:

$$\mathcal{L}(\boldsymbol{\theta}) = \lambda_{ode}\mathcal{L}(\boldsymbol{\theta})_{ode} + \lambda_{ic}\mathcal{L}(\boldsymbol{\theta})_{ic}$$

Consider a coupled mass–spring system with parameters $m_1 = 2$, $m_2 = 1$, $k_1 = 1$, and $k_2 = 1$, initial conditions $x(0) = 3$, $x'(0) = 0$ $y(0) = 3$, and $y'(0) = 0$ as described in the *Coupled Mass–Spring System*. Under these conditions, the general form of the *Coupled Mass–Spring System* reduces to:

$$2x'' + 2x - y = 0 \qquad y'' + (y-1)x = 0.$$

The analytical solution to this system is $x(t) = 2\cos t + \cos 2t$ and $y(t) = 4\cos t - \cos 2t$. **Figure 10(a)** illustrates the neural network architecture designed to solve the coupled mass–spring system, while **Figure 10(b)** compares the network predictions with the exact solution. **Figure 10(c)** shows the evolution of the total $L_2$ error (including the $L_2$ errors of $x$ and $y$) with respect to the training iterations, where the $L_2$ error is recorded every 500 iterations. The results demonstrate that, with appropriate hyperparameter tuning, the PINNs can accurately predict the system's behavior even in the absence of experimental data. The network configuration, hyperparameters, $L_2$ error, and computational time are summarized in **Table 5**.

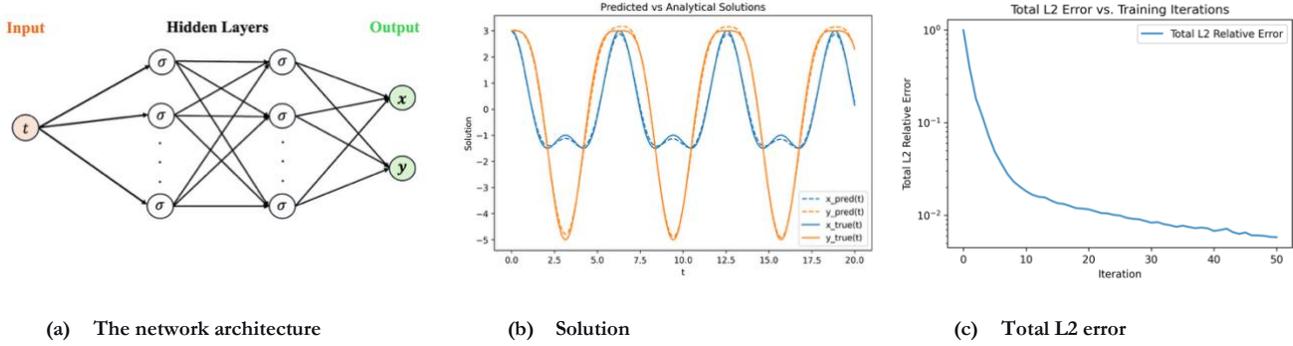

(a) The network architecture  (b) Solution  (c) Total L2 error

**Figure 10.** Graphical illustration of the network, solution and the L2 error.

| L | N | AF | Alg | R | C | E | LW | $L_2$ error | CT |
|---|---|---|---|---|---|---|---|---|---|
| 3 | 40 | tanh | Adams, L-BFGS | 0.001 | 400 | 50000 | [1;1] | 2.96e-03 | 220.51 |

**Table 5.** The network setting for the coupled mass–spring system. Abbreviations: L the number of layers; N the number of neurons; AF activation function; Alg optimization algorithm; R learning rate; C number of collocation points; E number of epochs; LW loss weights; CT computational time in seconds.

## CONCLUSIONS

In this paper, the Physics-Informed Neural Networks methodology for solving dynamical systems governed by ordinary differential equations is presented. Several strategies were introduced to enhance prediction accuracy, including residual weighting, the addition of feature layers, and the imposition of hard boundary constraints. The predictive capability of the PINNs approach was validated through four benchmark problems from engineering and biology, demonstrating that successful convergence for complex systems requires careful tuning of hyperparameters. The results highlight the strong predictive power of PINNs, establishing them as a promising tool for solving dynamical systems. A notable advantage of the PINNs framework is its flexibility, allowing inverse problems to be addressed with only minor adjustments to the implementation used for forward problems. Thus, future work will focus on extending the methodology to inverse problems, particularly for systems of nonlinear ODEs.

## ACKNOWLEDGEMENTS

The authors thank Dr. Charuka Wickramasinghe from Wayne State University and Prof. Pradeep Ranaweera from Siena Heights University for leading the experimental research.

**ABOUT THE STUDENT AUTHORS**

Andrew Particka is an undergraduate senior studying environmental engineering at Siena Heights University who will graduate in the spring of 2026.

Tyrus Whitman is an undergraduate student currently enrolled in his third year. He is looking to graduate in 2027 with a B.S in Mechanical Engineering.

Christopher Diers is a senior studying cybersecurity who will graduate from Siena Heights University in the spring of 2026.

Ian Griffin is a mechanical engineering major at Siena Heights University and is set to graduate in 2028.

**PRESS SUMMARY**

This study demonstrates and validates the predictive capability of Physics Informed Neural Networks for solving engineering and biological dynamical systems governed by ordinary differential equations. While traditional numerical methods often struggle with stiff, high-dimensional, or irregular problems, PINNs handle these challenges by embedding physical laws directly into the learning process. Balancing losses, tuning hyperparameters, and adding constraints without losing the generality of the ODE system are key to improving PINNs' accuracy and predictive power.